\documentclass[conference]{IEEEtran}
\IEEEoverridecommandlockouts
\usepackage{cite}
\usepackage{amsmath,amssymb,amsfonts}
\usepackage{algorithmic}
\usepackage{graphicx}
\usepackage{textcomp}
\usepackage{xcolor}
\usepackage{multicol}
\usepackage{multirow}
\usepackage{enumitem}
\usepackage{bbding}
\usepackage{url} 
\def\BibTeX{{\rm B\kern-.05em{\sc i\kern-.025em b}\kern-.08em
    T\kern-.1667em\lower.7ex\hbox{E}\kern-.125emX}}
\begin{document}

\title{TS-SAM: Fine-Tuning Segment-Anything Model for Downstream Tasks}

\author{\IEEEauthorblockN{1\textsuperscript{st} Yang Yu}
\IEEEauthorblockA{\textit{College of Computer Science} \\
\textit{NanKai University}\\
Tianjin, China \\
2120220639@mail.nankai.edu.cn}
\and
\IEEEauthorblockN{2\textsuperscript{nd} Chen Xu}
\IEEEauthorblockA{\textit{College of Computer Science} \\
\textit{NanKai University}\\
Tianjin, China \\
2110598@mail.nankai.edu.cn}
\and
\IEEEauthorblockN{3\textsuperscript{rd} Kai Wang$^{\ast}$ \thanks{*Corresponding author}}
\IEEEauthorblockA{\textit{College of Computer Science} \\
\textit{NanKai University}\\
Tianjin, China \\
wangk@nankai.edu.cn}
}

\maketitle

\begin{abstract}
Adapter based fine-tuning has been studied for improving the performance of SAM on downstream tasks.
However, there is still a significant performance gap between fine-tuned SAMs and domain-specific models.
To reduce the gap, we propose Two-Stream SAM (TS-SAM).
On the one hand, inspired by the side network in Parameter-Efficient Fine-Tuning (PEFT), we designed a lightweight Convolutional Side Adapter (CSA), which integrates the powerful features from SAM into side network training for comprehensive feature fusion.
On the other hand, in line with the characteristics of segmentation tasks, we designed Multi-scale Refinement Module (MRM) and Feature Fusion Decoder (FFD) to keep both the detailed and semantic features.
Extensive experiments on ten public datasets from three tasks demonstrate that TS-SAM not only significantly outperforms the recently proposed SAM-Adapter and SSOM, but achieves competitive performance with the SOTA domain-specific models.
Our code is available at: \url{https://github.com/maoyangou147/TS-SAM}.
\end{abstract}


\section{Introduction}

As a large vision model pretrained on over 11 million images, Segment-Anything Model (SAM)\cite{c3} has attracted the researchers' interests.
However, recent studies demonstrate that SAM struggles to achieve satisfactory performance on downstream tasks, including Camouflaged Object Detection (COD)\cite{c17,c4}, shadow detection\cite{c4} and Salient Object Detection (SOD)\cite{c5}.
\par
It is a critical issue in the application of large models that how to better adapt large models, pretrained on massive, general-purpose datasets, to different downstream tasks.
To address this issue, there already exist numerous studies for Parameter-Efficient Fine-Tuning (PEFT)\cite{c8,c9,c11}. Some methods\cite{c9,c11} employ lightweight Adapters or Prompts to bridge the gap between general-purpose large models and various downstream tasks.
Only a small number of Adapter or Prompt parameters are updated during training, reducing storage and computational costs.
Recently, fine-tuning methods based on side networks\cite{c8,c14,c16} have also gained attention.
This approach adds a lightweight side network to the large model, fine-tuning only the side network during training. 
\begin{figure}[htbp]
\centering
\includegraphics[width=0.9\linewidth]{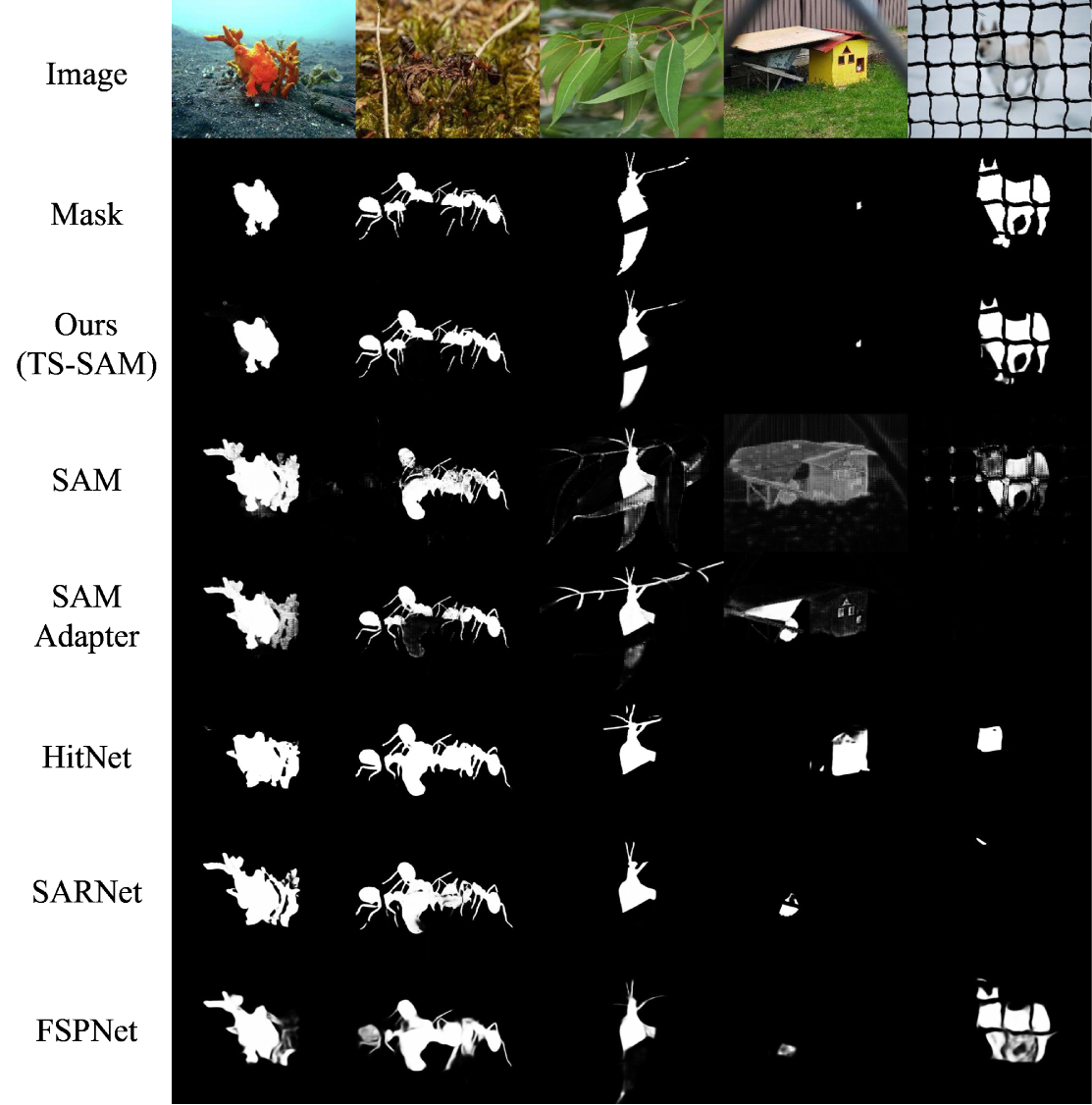}
\caption{Comparison of TS-SAM with SAM, SAM-Adapter and the SOTA domain-specific models on some images from COD10K dataset.}
\label{fig:compare_on_cod}
\end{figure}
The side network adapts the features extracted by the large model's backbone to the requirements of downstream tasks.\par
Currently, there have been some efforts to fine-tune SAM using PEFT.
SAM-Adapter\cite{c4} introduces lightweight adapters into the SAM encoder, which improves the performance of SAM on COD and shadow detection tasks.
SSOM\cite{c5} adaptively fine-tunes SAM using the inherent low-rank structure, which improves the performance of SAM on SOD task.
Both SAM-Adapter and SSOM are pioneering works in exploring the ability of SAM to be applied to downstream tasks.
However, there is still a significant performance gap between these fine-tuned SAMs and recent domain-specific models.

In this paper, we aim to address the challenge of SAM's suboptimal performance in various downstream tasks by devising a unified fine-tuning strategy to boost SAM's efficacy across diverse application scenarios.
Inspired by the fine-tuning methods based on side networks, we propose Two-Stream SAM (TS-SAM) to uniformly fine-tune SAM across different downstream tasks.
Specifically, we designed a lightweight Convolutional Side Adapter (CSA) to assist SAM in operating under various challenging scenarios.
In addition, in line with the characteristics of segmentation tasks, we proposed the Multi-scale Refinement Module (MRM) to extract finer positional features of images for more granular segmentation.
For the decoding process, we designed a Feature Fusion Decoder (FFD) to integrate features of different scales during decoding, yielding refined segmentation results.
Fig. \ref{fig:compare_on_cod} shows the comparison of the proposed TS-SAM with SAM, SAM-Adapter and the SOTA domain-specific models on some images from COD10K dataset, demonstrating the superiority of TS-SAM.
Also, TS-SAM is lightweight, with the ViT-h version requiring only 29.44M trainable parameters, constituting 4.4\% of the total model parameter count.
This allows for the storage of only a small number of parameter copies for different downstream tasks.\par
The contributions of this work are summarized as follows:
\begin{enumerate}
\setlength{\leftmargin}{0pt}
    \item We introduced the side network into the fine-tuning of SAM for the first time. Innovatively, we proposed the structure of a two-stream side network, which effectively extracts features from the SAM encoder.
    \item We proposed the Multi-Scale Refinement Module (MRM) and the Feature Fusion Decoder (FFD) tailored for segmentation tasks. These modules acquire refined target positional information through high-resolution hierarchical features and fully integrate them during the decoding process to achieve detailed segmentation results.
    \item We evaluated our proposed TS-SAM on ten public datasets from three tasks, including COD, shadow detection and SOD. The experimental results demonstrate that TS-SAM significantly outperforms recent works that fine-tune SAM for these downstream tasks. It even achieves competitive performance compared to the SOTA domain-specific models specifically designed for each task.
\end{enumerate}
\section{Proposed Methods}
\label{sec:methods}
\begin{figure*}[htbp]
\centering
\includegraphics[width=0.8\textwidth]{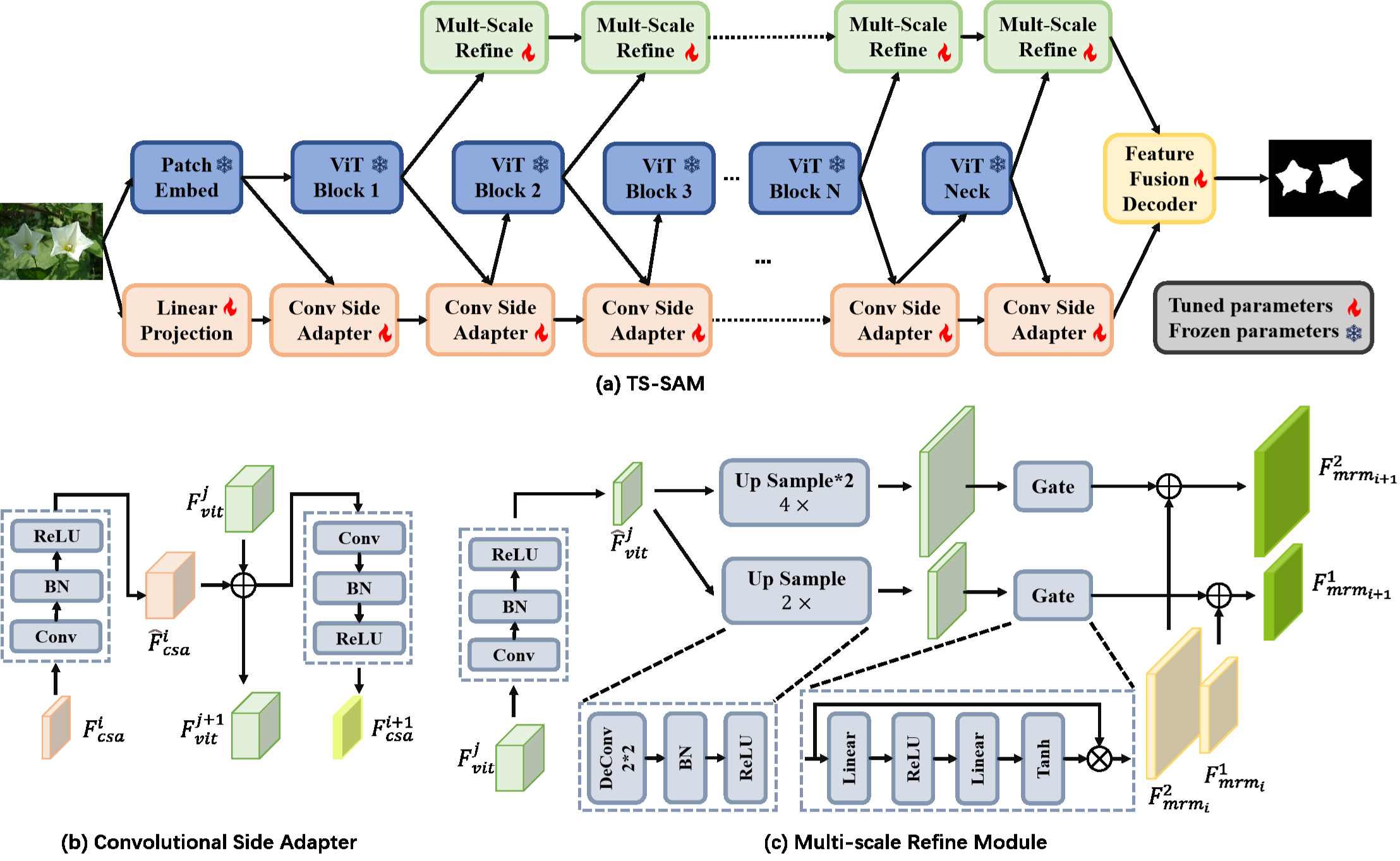}
\caption{(a) Overall architecture of TS-SAM. (b) a Convolutional Side Adapter (CSA) for extracting visual features from the SAM image encoder and adapting them to downstream tasks. (c) a Multi-scale Refinement Module (MRM) for extracting detailed features from images.}
\label{fig:model_structure}
\end{figure*}

\subsection{Overall Architecture} \par
Fig. \ref{fig:model_structure}(a) illustrates the overall architecture of the proposed TS-SAM.
We used the pretrained SAM ViT as the backbone network and designed a series of lightweight modules for fine-tuning in downstream tasks.
Given an image $I\in\mathbb{R}^{3*H*W}$, visual features $F_{vit}\in\mathbb{R}^{C*\frac{H}{16}*\frac{W}{16}}$ are extracted through the SAM image encoder.
Simultaneously, features from the SAM image encoder are extracted layer by layer through stacked Convolutional Side Adapters (CSA), resulting in image features $F_{csa}\in\mathbb{R}^{C_1*\frac{H}{16}*\frac{W}{16}}$ adapted to downstream tasks.
To extract more detailed features from the image encoder, we propose the Multi-scale Refinement Module (MRM).
MRM upsamples the feature embeddings from various layers of the image encoder, creating a hierarchical feature representation ${\{F_{mrm}^{k}\}}_{k=1}^{2}$.
Further, a lightweight gating unit continuously merges features from lower to higher layers of the SAM image encoder, gathering a richer array of image detail features.\par
During the decoding process, we do not use the SAM mask decoder.
The reason is that the SAM decoder requires prompts such as points or boxes to achieve good results, and it is challenging to segment multiple targets through a single forward pass.
Therefore, we designed the lightweight Feature Fusion Decoder (FFD) to inject the hierarchical feature representation ${\{F_{mrm}^{k}\}}_{k=1}^{2}$ into the features $F_{csa}$ obtained from CSA.
This enhances the feature representation, resulting in a refined segmentation mask.
The FFD gradually merges the hierarchical feature representation during the upsampling process on $F_{csa}$, highlighting key information in the hierarchical feature representation through a two-stage injection approach.\par

Finally, in order to reduce training costs, the SAM image encoder is frozen during the training process.
Only Convolutional Side Adapter, Multi-Scale Refinement Module, and Feature Fusion Decoder are trained, and all three components are lightweight.

\subsection{Convolutional Side Adapter} \par
This subsection provides detailed introduction to the Convolutional Side Adapter (CSA).
Inspired by adapter techniques in the field of PEFT, we have made simple modifications to the original adapter.
We believe that CSA can effectively extract features from the SAM image encoder and bridge the gap between these features and the data of downstream tasks.
As shown in Fig. \ref{fig:model_structure}(b), unlike the original adapter, CSA comprises two $1*1$ convolution modules. The first $1*1$ convolution expands the compressed features to the feature dimension of the SAM image encoder, and then merges with the output feature embeddings of the SAM image encoder.
The second $1*1$ convolution compresses the merged features back to the input feature dimension, serving as the input for the next CSA layer. Given the output features $F_{csa}^{i}\in\mathbb{R}^{C_1*\frac{H}{16}*\frac{W}{16}}$ from the (i-1)-th CSA and the output features $F_{vit}^{j}\in\mathbb{R}^{C*\frac{H}{16}*\frac{W}{16}}$ from the (j-1)-th layer of the SAM image encoder, the operation of the i-th CSA module can be represented as follows:
\begin{equation}
F_{vit}^{j+1}=F_{vit}^{j}+{conv}_{1*1}(F_{csa}^{i})  \tag{1}
\end{equation}
\begin{equation}
F_{csa}^{i+1}={conv}_{1*1}(F_{vit}^{j+1})  \tag{2}
\end{equation}
where ${conv}_{1*1}$ represents the $1*1$ convolution module, which includes a $1*1$ convolution, batch normalization and activation operations. $F_{vit}^{j+1}$ is the input to the j-th layer of the SAM image encoder, and $F_{csa}^{i+1}$ is the input to the ith CSA.
CSA is lightweight, maintaining the simplicity of the adapter. 

\subsection{Multi-Scale Refine Module} \par
In segmentation tasks, to achieve more precise segmentation results, a model needs the ability to effectively describe detailed features such as object edges. However, the 16x downsampling of images during the patch embedding in the SAM image encoder might result in difficulty extracting target positional information. Therefore, we propose the Multi-Scale Refinement Module (MRM) to obtain higher-resolution features with more details.\par

Fig. \ref{fig:model_structure}(c) illustrates the structure of the MRM.
The i-th MRM layer receive input of output features $F_{vit}^{j}$ from the (j-1)-th layer of the SAM image encoder and the hierarchical features ${\{F_{mrm_{i}}^{k}\}}_{k=1}^{2}$ from the (i-1)-th layer of the MRM, output two features ${\{F_{mrm_{i+1}}^{k}\}}_{k=1}^{2}$ which resolutions are $\frac{H}{8} \times \frac{W}{8}$ and $\frac{H}{4} \times \frac{W}{4}$. 
We first compress the feature dimensions of $F_{vit}^{j}$ through a $1*1$ convolution module to obtain ${\widehat{F}}_{vit}^{j}\in\mathbb{R}^{C^{'}*\frac{H}{16}*\frac{W}{16}}$.
Then, ${\widehat{F}}_{vit}^{j}$ is processed through a deconvolution module to yield a higher-resolution hierarchical feature representation ${\{{\widehat{F}}_{mrm_{i+1}}^{k}\}}_{k=1}^{2}$. The above process can be formalized as follows:
\begin{equation}
{\widehat{F}}_{mrm_{i+1}}^{k}={deconv}_k({conv}_{1*1}(F_{vit}^{j})),k=1,2  \tag{3}
\end{equation}
where $deconv$ represents the deconvolution module, which performs 2x and 4x upsampling on ${\widehat{F}}_{vit}^{j}$, resulting in high-resolution features at two different scales.
Further, the obtained high-resolution features need to be merged with the hierarchical features output by the previous layer of the MRM.
To control the extent of feature fusion, we employed the lightweight gating unit proposed in \cite{c20} for the high-resolution features at different scales.
This unit calculates pixel-level weights through linear layers and activation operations, thereby finely controlling the extent of feature fusion.
The operation of the gating unit can be formalized as follows:
\begin{equation}
\resizebox{.9\hsize}{!}{${{\widetilde{F}}_{mrm_{i+1}}^{k}=Tanh(Linear(ReLU(Linear({\widehat{F}}_{{mrm}_{i+1}}^{k}))))\otimes {\widehat{F}}_{{mrm}_{i+1}}^{k},k=1,2}$}\tag{4}
\end{equation}
where $Linear$ represents the linear layer, and $\otimes$ denotes element-wise multiplication.
Finally, the two features are simply summed to achieve feature fusion:
\begin{equation}
F_{{mrm}_{i+1}}^{k}={\widetilde{F}}_{mrm_{i+1}}^{k}+F_{{mrm}_i}^{k},k=1,2  \tag{5}
\end{equation}
\renewcommand{\floatpagefraction}{.9}
\begin{figure}[t]
\centering
\includegraphics[width=0.7\linewidth]{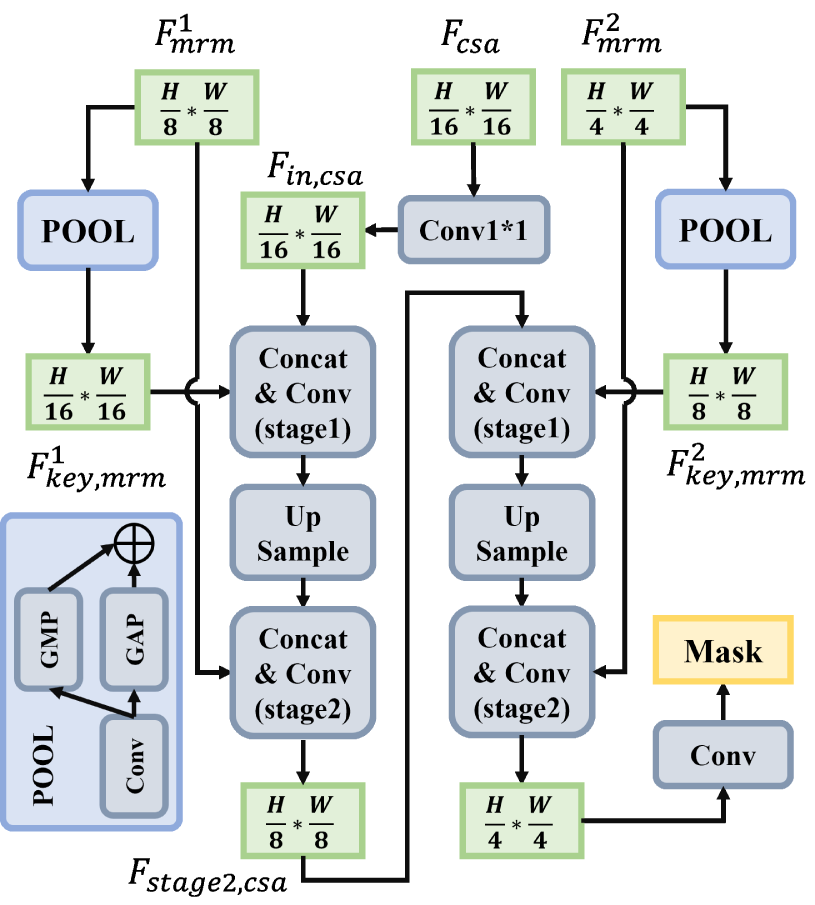}
\caption{Structure of Feature Fusion Decoder (FFD), which injects $F_{mrm}^{1}$ and $F_{mrm}^{2}$ respectively into $F_{csa}$. Rectangular boxes represent feature maps of different scales, while rounded boxes represent different modules.}

\label{fig:ffd_structure}
\end{figure}
\vspace{-2em}
\subsection{Feature Fusion Decoder} \par
To ensure the thorough integration of features from the CSA and MRM branches, we propose a lightweight Feature Fusion Decoder (FFD).
FFD is designed to inject hierarchical feature representations into the features obtained from the CSA branch during the decoding process.
The structure of FFD is shown in Fig. \ref{fig:ffd_structure}.
CSA branch feature $F_{csa}$ is first processed through a $1*1$ convolution to get $F_{in,csa}$.
For the hierarchical feature representation ${\{F_{mrm}^{k}\}_{k=1}^{2}}$ whose resolutions are $\frac{H}{8}*\frac{W}{8}$ and $\frac{H}{4}*\frac{W}{4}$, respectively, we aim to highlight its key features. Therefore, we employ a pooling operation to downsample the hierarchical feature representation, emphasizing its crucial components. This process can be represented as follows:

\begin{equation}
F_{in,mrm}^{k}={conv}_{1*1}(F_{mrm}^{k}),k=1,2  \tag{6}
\end{equation}
\begin{equation}
\resizebox{.9\hsize}{!}{${F_{key,mrm}^{k}=GAP(F_{in,mrm}^{k})+GMP(F_{in,mrm}^{k}),k=1,2}$}  \tag{7}
\end{equation}
where $GAP$ (Global Average Pooling) and $GMP$ (Global Max Pooling) respectively have 2*2 kernel. For high-resolution features at different scales, we adopt a two-stage injection approach. First, the key features $F_{key,mrm}$ obtained through the pooling operation are injected into $F_{in,csa}$, and then the complete high-resolution hierarchical feature $F_{mrm}$ are injected into $F_{in,csa}$. Taking the injection process of the high-resolution feature $F_{mrm}^{1}$, having the resolution of $\frac{H}{8}*\frac{W}{8}$, as an example, the fusion operation can be formalized as follows:
\begin{equation}
F_{stage1,csa}={conv}_{3*3}([F_{in,csa},F_{key,mrm}^{1}])  \tag{8}
\end{equation}
\begin{equation}
{\widehat{F}}_{stage1,csa}=UpSample(F_{stage1,csa})  \tag{9}
\end{equation}
\begin{equation}
F_{stage2,csa}={conv}_{3*3}([{\widehat{F}}_{stage1,csa},F_{mrm}^{1}])  \tag{10}
\end{equation}

where Equation (8) and Equation (10) respectively represent the first and second stages of the injection process, with denoting concatenation along the channel dimension, and $UpSample$ representing 2x upsampling.
Following this, the injection process for the high-resolution feature $F_{mrm}^{2}$, having the resolution of $\frac{H}{4}*\frac{W}{4}$, is identical to the process above.
Through this two-stage injection, the hierarchical features are fully integrated with the features from the CSA branch, resulting in an enhanced feature representation.
This allows the decoding process to achieve more refined segmentation results.

\begin{table*}[!htbp]
\centering
\caption{Quantitative comparison of TS-SAM with SAM, SAM-Adapter and the SOTA domain-specific models on four datasets of the COD task: CHAMELEON, CAMO, COD10K, and NC4K. The best results are marked in \textbf{bold}.}\label{tab1}
\scalebox{0.8}{
\begin{tabular}{ccccccccccccccccc}
\hline
\multicolumn{1}{c||}{\multirow{2}{*}{Methods}} 
& \multicolumn{4}{c|}{\textbf{CHAMELEON}}
& \multicolumn{4}{c|}{\textbf{CAMO}}
& \multicolumn{4}{c|}{\textbf{COD10K}}
& \multicolumn{4}{c}{\textbf{NC4K}} \\ \cline{2-17}

\multicolumn{1}{c||}{}                
& $S_\alpha \uparrow$ & $E_\phi\uparrow$ 
    & $F_\beta^\omega\uparrow$ & \multicolumn{1}{c|}{$MAE\downarrow$}
& $S_\alpha \uparrow$ & $E_\phi\uparrow$ 
    & $F_\beta^\omega\uparrow$ & \multicolumn{1}{c|}{$MAE\downarrow$}
& $S_\alpha \uparrow$ & $E_\phi\uparrow$ 
    & $F_\beta^\omega\uparrow$ &\multicolumn{1}{c|}{$MAE\downarrow$}
& $S_\alpha \uparrow$ & $E_\phi\uparrow$ 
    & $F_\beta^\omega\uparrow$ &\multicolumn{1}{c}{$MAE\downarrow$}
\\ \hline

\multicolumn{1}{c||}{$\rm FDCOD_{22}$\cite{fdcod}}
    & \multicolumn{1}{c}{0.834} & \multicolumn{1}{c}{0.893} 
    & \multicolumn{1}{c}{0.750} & \multicolumn{1}{c|}{0.051}   
    & \multicolumn{1}{c}{0.828} & \multicolumn{1}{c}{0.883} 
    & \multicolumn{1}{c}{0.748} & \multicolumn{1}{c|}{0.068} 
    & \multicolumn{1}{c}{0.832} & \multicolumn{1}{c}{0.907} 
    & \multicolumn{1}{c}{0.706} & \multicolumn{1}{c|}{0.033}  
    & \multicolumn{1}{c}{0.834} & \multicolumn{1}{c}{0.893} 
    & \multicolumn{1}{c}{0.750} & \multicolumn{1}{c}{0.051}  
    \\ 
    
\multicolumn{1}{c||}{$\rm SegMaR_{22}$\cite{segmar}}
    & \multicolumn{1}{c}{0.841} & \multicolumn{1}{c}{0.896} 
    & \multicolumn{1}{c}{0.781} & \multicolumn{1}{c|}{0.046} 
    & \multicolumn{1}{c}{0.815} & \multicolumn{1}{c}{0.874} 
    & \multicolumn{1}{c}{0.753} & \multicolumn{1}{c|}{0.071} 
    & \multicolumn{1}{c}{0.833} & \multicolumn{1}{c}{0.899} 
    & \multicolumn{1}{c}{0.724} & \multicolumn{1}{c|}{0.034}  
    & \multicolumn{1}{c}{0.841} & \multicolumn{1}{c}{0.896} 
    & \multicolumn{1}{c}{0.781} & \multicolumn{1}{c}{0.046} 
    \\
    
\multicolumn{1}{c||}{$\rm ZoomNet_{22}$\cite{zoomnet}}
    & \multicolumn{1}{c}{0.853} & \multicolumn{1}{c}{0.896} 
    & \multicolumn{1}{c}{0.784} & \multicolumn{1}{c|}{0.043} 
    & \multicolumn{1}{c}{0.820} & \multicolumn{1}{c}{0.877} 
    & \multicolumn{1}{c}{0.752} & \multicolumn{1}{c|}{0.066} 
    & \multicolumn{1}{c}{0.838} & \multicolumn{1}{c}{0.888} 
    & \multicolumn{1}{c}{0.729} & \multicolumn{1}{c|}{0.029}  
    & \multicolumn{1}{c}{0.853} & \multicolumn{1}{c}{0.896} 
    & \multicolumn{1}{c}{0.784} & \multicolumn{1}{c}{0.043}  
    \\ 
    
\multicolumn{1}{c||}{$\rm SINetV2_{22}$\cite{sinetv2}}
    & \multicolumn{1}{c}{0.847} & \multicolumn{1}{c}{0.903} 
    & \multicolumn{1}{c}{0.770} & \multicolumn{1}{c|}{0.048} 
    & \multicolumn{1}{c}{0.822} & \multicolumn{1}{c}{0.882} 
    & \multicolumn{1}{c}{0.743} & \multicolumn{1}{c|}{0.070} 
    & \multicolumn{1}{c}{0.815} & \multicolumn{1}{c}{0.887}
    & \multicolumn{1}{c}{0.680} & \multicolumn{1}{c|}{0.037}  
    & \multicolumn{1}{c}{0.847} & \multicolumn{1}{c}{0.903}
    & \multicolumn{1}{c}{0.770} & \multicolumn{1}{c}{0.048}  
    \\
    
\multicolumn{1}{c||}{$\rm FAPNet_{22}$\cite{fapnet}}
    & \multicolumn{1}{c}{0.893} & \multicolumn{1}{c}{0.940} 
    & \multicolumn{1}{c}{-} & \multicolumn{1}{c|}{0.028}  
    & \multicolumn{1}{c}{0.815} & \multicolumn{1}{c}{0.865} 
    & \multicolumn{1}{c}{0.734} & \multicolumn{1}{c|}{0.076}
    & \multicolumn{1}{c}{0.822} & \multicolumn{1}{c}{0.887} 
    & \multicolumn{1}{c}{0.694} & \multicolumn{1}{c|}{0.036}
    & \multicolumn{1}{c}{0.851} & \multicolumn{1}{c}{0.899} 
    & \multicolumn{1}{c}{0.775} & \multicolumn{1}{c}{0.046}
    \\
    
\multicolumn{1}{c||}{$\rm ICON_{22}$\cite{icon}}
    & \multicolumn{1}{c}{0.847} & \multicolumn{1}{c}{0.911}
    & \multicolumn{1}{c}{0.784} & \multicolumn{1}{c|}{0.045}  
    & \multicolumn{1}{c}{0.840} & \multicolumn{1}{c}{0.894} 
    & \multicolumn{1}{c}{0.769} & \multicolumn{1}{c|}{0.058} 
    & \multicolumn{1}{c}{0.818} & \multicolumn{1}{c}{0.904} 
    & \multicolumn{1}{c}{0.688} & \multicolumn{1}{c|}{0.033} 
    & \multicolumn{1}{c}{0.847} & \multicolumn{1}{c}{0.911} 
    & \multicolumn{1}{c}{0.784} & \multicolumn{1}{c}{0.045} 
    \\

\multicolumn{1}{c||}{$\rm FEDER_{23}$\cite{feder}}
    & \multicolumn{1}{c}{0.887} & \multicolumn{1}{c}{0.954}
    & \multicolumn{1}{c}{0.835} & \multicolumn{1}{c|}{0.030}  
    & \multicolumn{1}{c}{0.802} & \multicolumn{1}{c}{0.873} 
    & \multicolumn{1}{c}{0.738} & \multicolumn{1}{c|}{0.071} 
    & \multicolumn{1}{c}{0.822} & \multicolumn{1}{c}{0.905} 
    & \multicolumn{1}{c}{0.716} & \multicolumn{1}{c|}{0.032} 
    & \multicolumn{1}{c}{0.847} & \multicolumn{1}{c}{0.915} 
    & \multicolumn{1}{c}{0.789} & \multicolumn{1}{c}{0.044} 
    \\

\multicolumn{1}{c||}{$\rm DGNet_{23}$\cite{dgnet}}
    & \multicolumn{1}{c}{0.890} & \multicolumn{1}{c}{0.956}
    & \multicolumn{1}{c}{0.816} & \multicolumn{1}{c|}{0.029}  
    & \multicolumn{1}{c}{0.838} & \multicolumn{1}{c}{0.814} 
    & \multicolumn{1}{c}{0.768} & \multicolumn{1}{c|}{0.057} 
    & \multicolumn{1}{c}{0.822} & \multicolumn{1}{c}{0.911} 
    & \multicolumn{1}{c}{0.692} & \multicolumn{1}{c|}{0.033} 
    & \multicolumn{1}{c}{0.857} & \multicolumn{1}{c}{0.922} 
    & \multicolumn{1}{c}{0.783} & \multicolumn{1}{c}{0.042} 
    \\

\multicolumn{1}{c||}{$\rm SARNet_{23}$\cite{sarnet}}
    & \multicolumn{1}{c}{0.933} & \multicolumn{1}{c}{\textbf{0.978}}
    & \multicolumn{1}{c}{\textbf{0.909}} & \multicolumn{1}{c|}{\textbf{0.017}}  
    & \multicolumn{1}{c}{0.874} & \multicolumn{1}{c}{0.935} 
    & \multicolumn{1}{c}{\textbf{0.844}} & \multicolumn{1}{c|}{\textbf{0.046}} 
    & \multicolumn{1}{c}{0.885} & \multicolumn{1}{c}{0.947} 
    & \multicolumn{1}{c}{0.820} & \multicolumn{1}{c|}{0.021} 
    & \multicolumn{1}{c}{0.889} & \multicolumn{1}{c}{0.940} 
    & \multicolumn{1}{c}{0.851} & \multicolumn{1}{c}{\textbf{0.032}} 
    \\

\multicolumn{1}{c||}{$\rm FSPNet_{23}$\cite{fspnet}}
    & \multicolumn{1}{c}{0.908} & \multicolumn{1}{c}{0.965}
    & \multicolumn{1}{c}{0.851} & \multicolumn{1}{c|}{0.023}  
    & \multicolumn{1}{c}{0.856} & \multicolumn{1}{c}{0.928} 
    & \multicolumn{1}{c}{0.799} & \multicolumn{1}{c|}{0.050} 
    & \multicolumn{1}{c}{0.851} & \multicolumn{1}{c}{0.930} 
    & \multicolumn{1}{c}{0.735} & \multicolumn{1}{c|}{0.026} 
    & \multicolumn{1}{c}{0.878} & \multicolumn{1}{c}{0.937} 
    & \multicolumn{1}{c}{0.816} & \multicolumn{1}{c}{0.035} 
    \\

\multicolumn{1}{c||}{$\rm MSCAF-Net_{23}$\cite{mscafnet}}
    & \multicolumn{1}{c}{0.912} & \multicolumn{1}{c}{0.970}
    & \multicolumn{1}{c}{0.865} & \multicolumn{1}{c|}{0.022}  
    & \multicolumn{1}{c}{0.873} & \multicolumn{1}{c}{\textbf{0.937}} 
    & \multicolumn{1}{c}{0.828} & \multicolumn{1}{c|}{\textbf{0.046}} 
    & \multicolumn{1}{c}{0.865} & \multicolumn{1}{c}{0.936} 
    & \multicolumn{1}{c}{0.776} & \multicolumn{1}{c|}{0.024} 
    & \multicolumn{1}{c}{0.887} & \multicolumn{1}{c}{\textbf{0.942}} 
    & \multicolumn{1}{c}{0.838} & \multicolumn{1}{c}{\textbf{0.032}} 
    \\

\multicolumn{1}{c||}{$\rm HitNet_{23}$\cite{hitnet}}
    & \multicolumn{1}{c}{0.921} & \multicolumn{1}{c}{0.972}
    & \multicolumn{1}{c}{0.897} & \multicolumn{1}{c|}{0.019}  
    & \multicolumn{1}{c}{0.849} & \multicolumn{1}{c}{0.910} 
    & \multicolumn{1}{c}{0.809} & \multicolumn{1}{c|}{0.055} 
    & \multicolumn{1}{c}{0.871} & \multicolumn{1}{c}{0.938} 
    & \multicolumn{1}{c}{0.806} & \multicolumn{1}{c|}{0.023} 
    & \multicolumn{1}{c}{0.875} & \multicolumn{1}{c}{0.929} 
    & \multicolumn{1}{c}{0.834} & \multicolumn{1}{c}{0.037} 
    \\ \hline
    
\multicolumn{1}{c||}{$\rm SAM_{23}$\cite{c3}} 
    & \multicolumn{1}{c}{0.767} & \multicolumn{1}{c}{0.776} 
    & \multicolumn{1}{c}{0.696} & \multicolumn{1}{c|}{0.078}  
    & \multicolumn{1}{c}{0.684} & \multicolumn{1}{c}{0.687} 
    & \multicolumn{1}{c}{0.606} & \multicolumn{1}{c|}{0.132} 
    & \multicolumn{1}{c}{0.783} & \multicolumn{1}{c}{0.798} 
    & \multicolumn{1}{c}{0.701} & \multicolumn{1}{c|}{0.050} 
    & \multicolumn{1}{c}{0.767} & \multicolumn{1}{c}{0.776} 
    & \multicolumn{1}{c}{0.696} & \multicolumn{1}{c}{0.078} 
    \\
    
\multicolumn{1}{c||}{$\rm SAM-Adapter_{23}$\cite{c4}}
    & \multicolumn{1}{c}{0.896} & \multicolumn{1}{c}{0.919} 
    & \multicolumn{1}{c}{0.824} & \multicolumn{1}{c|}{0.033} 
    & \multicolumn{1}{c}{0.847} & \multicolumn{1}{c}{0.873} 
    & \multicolumn{1}{c}{0.765} & \multicolumn{1}{c|}{0.070} 
    & \multicolumn{1}{c}{0.883} & \multicolumn{1}{c}{0.918} 
    & \multicolumn{1}{c}{0.801} & \multicolumn{1}{c|}{0.025} 
    & \multicolumn{1}{c}{-} & \multicolumn{1}{c}{-} 
    & \multicolumn{1}{c}{-} & \multicolumn{1}{c}{-} 
    \\

\multicolumn{1}{c||}{$\rm TS-SAM\_B(Ours)$} 
    & \multicolumn{1}{c}{0.912} & \multicolumn{1}{c}{0.947} 
    & \multicolumn{1}{c}{0.849} & \multicolumn{1}{c|}{0.023} 
    & \multicolumn{1}{c}{0.826} & \multicolumn{1}{c}{0.862} 
    & \multicolumn{1}{c}{0.753} & \multicolumn{1}{c|}{0.073} 
    & \multicolumn{1}{c}{0.863} & \multicolumn{1}{c}{0.909} 
    & \multicolumn{1}{c}{0.771} & \multicolumn{1}{c|}{0.029} 
    & \multicolumn{1}{c}{0.866} & \multicolumn{1}{c}{0.900} 
    & \multicolumn{1}{c}{0.799} & \multicolumn{1}{c}{0.045} 
    \\

\multicolumn{1}{c||}{$\rm TS-SAM\_H(Ours)$} 
    & \multicolumn{1}{c}{\textbf{0.942}} & \multicolumn{1}{c}{0.966} 
    & \multicolumn{1}{c}{0.900} & \multicolumn{1}{c|}{\textbf{0.017}} 
    & \multicolumn{1}{c}{\textbf{0.887}} & \multicolumn{1}{c}{0.925} 
    & \multicolumn{1}{c}{0.830} & \multicolumn{1}{c|}{0.047}
    & \multicolumn{1}{c}{\textbf{0.914}} & \multicolumn{1}{c}{\textbf{0.950}} 
    & \multicolumn{1}{c}{\textbf{0.856}} & \multicolumn{1}{c|}{\textbf{0.017}}  
    & \multicolumn{1}{c}{\textbf{0.905}} & \multicolumn{1}{c}{0.933} 
    & \multicolumn{1}{c}{\textbf{0.860}} & \multicolumn{1}{c}{\textbf{0.032}}  
    
\\ \hline
% \hline

% \multicolumn{1}{c||}{Rank}
%     & \multicolumn{1}{c}{\textbf{1}} & \multicolumn{1}{c}{\textbf{4}} 
%     & \multicolumn{1}{c}{\textbf{2}} & \multicolumn{1}{c|}{\textbf{1}} 
%     & \multicolumn{1}{c}{\textbf{1}} & \multicolumn{1}{c}{\textbf{4}} 
%     & \multicolumn{1}{c}{\textbf{2}} & \multicolumn{1}{c|}{\textbf{3}} 
%     & \multicolumn{1}{c}{\textbf{1}} & \multicolumn{1}{c}{\textbf{1}} 
%     & \multicolumn{1}{c}{\textbf{1}} & \multicolumn{1}{c|}{\textbf{1}}
%     & \multicolumn{1}{c}{\textbf{1}} & \multicolumn{1}{c}{\textbf{4}} 
%     & \multicolumn{1}{c}{\textbf{1}} & \multicolumn{1}{c}{\textbf{1}}
%     \\ \hline
\end{tabular}
}
\end{table*}

\section{Experiment}
\label{sec:exp}

\subsection{Datasets and Implementation}
\noindent{\textbf{Datasets.}} 
In this subsection, we report the experiments on ten datasets from three challenging downstream tasks used in SAM-Adapter\cite{c4} and SSOM\cite{c5}.

%to validate the effectiveness of the proposed method.
% And experiments on another five datasets for the SOD task are introduced in \textbf{\textit{supplementary materials}}.
\par
\begin{enumerate}
    \setlength{\leftmargin}{0pt}
    \item COD: Four commonly used datasets: CAMO\cite{camo}, COD10K\cite{cod10k1}, CHAMELEON\cite{chameleon}, and NC4K\cite{nc4k}, were used to test the performance of TS-SAM. Following the training protocol in \cite{c4}, the training samples of \cite{camo} and \cite{cod10k1} were combined for the model training, and the testing samples in each dataset were used for testing.
    \item Shadow detection: ISTD\cite{istd} dataset was used to test the performance of TS-SAM.
    \item SOD: We selected five datasets: DUTS\cite{duts}, ECSSD\cite{ecssd}, OMRON\cite{omron}, HKU-IS\cite{hkuis}, and PASCAL-S\cite{pascals}.We followed the training protocol in \cite{c21}, using the DUTS training set for training and testing on the remaining datasets.
\end{enumerate}

\noindent{\textbf{Implementation Details.} The proposed solution is implemented in PyTorch, and all training was carried out using 4 NVIDIA A40 GPUs.
We trained two versions of the model: TS-SAM\_B and TS-SAM\_H.
TS-SAM\_B uses the ViT-B version of the SAM image encoder and includes 14 layers of CSA and 13 layers of MRM.
TS-SAM\_H uses the ViT-H version of the SAM image encoder and includes 34 layers of CSA and 13 layers of MRM.
All experiments uses the Adam optimization algorithm, with an initial learning rate set at 0.0008, and employed a cosine decay strategy.
The total batch size was 8.
No data augmentation methods were used besides Resize.
For the COD and SOD task, Binary Cross Entropy (BCE) loss and IOU loss are used.
Training was carried out for 80 epochs. For the shadow detection task, Balanced Binary Cross Entropy (BBCE) loss is used. Training was carried out for 100 epochs.
Specifically, we introduced the high frequency component of the image as input in our experiments on shadow detection task, following the same setting in \cite{c4}.\par
\begin{table}[htbp]
\centering
\caption{Quantitative comparison of TS-SAM with SAM, SAM-Adapter and the SOTA domain-specific models on the ISTD dataset of the shadow detection task. The best result is marked in \textbf{bold}.}\label{tab2}
\scalebox{0.9}{
\begin{tabular}{cc}
\hline
%    \multicolumn{1}{c||}{\multirow{2}{*}{Methods}} 
%    & \multicolumn{1}{c}{\textbf{ISTD}} \\ \cline{2-2}
    \multicolumn{1}{c||}{Methods}
    %\multicolumn{1}{c||}{} & 
    &\multicolumn{1}{c}{$BER\downarrow$}
    \\ \hline
    \multicolumn{1}{c||}{$\rm SDCM_{22}$\cite{sdcm}} & \multicolumn{1}{c}{1.41}\\
    \multicolumn{1}{c||}{$\rm TransShadow_{22}$\cite{transshadow}} 
        & \multicolumn{1}{c}{1.73}\\
    \multicolumn{1}{c||}{$\rm FCSDNet_{23}$\cite{fcsdnet}} & \multicolumn{1}{c}{1.69}\\
    \multicolumn{1}{c||}{$\rm RMLANet_{23}$\cite{rmlanet}} 
        & \multicolumn{1}{c}{\textbf{1.01}}\\
    \multicolumn{1}{c||}{$\rm SDDNet_{23}$\cite{sddnet}} & \multicolumn{1}{c}{1.27}\\
    \multicolumn{1}{c||}{$\rm SARA_{23}$\cite{sara}} & \multicolumn{1}{c}{1.18}
    \\ \hline
    \multicolumn{1}{c||}{$\rm SAM_{23}$\cite{c3}} & \multicolumn{1}{c}{40.51}\\
    \multicolumn{1}{c||}{$\rm SAM-Adapter_{23}$\cite{c4}} 
        & \multicolumn{1}{c}{1.43} \\
    \multicolumn{1}{c||}{$\rm TS-SAM\_B(Ours)$} 
    & \multicolumn{1}{c}{1.11} \\
    \multicolumn{1}{c||}{$\rm TS-SAM\_H(Ours)$} 
    & \multicolumn{1}{c}{1.04} \\ \hline
\end{tabular}
}
\end{table}
\subsection{Results}
\noindent{\textbf{Camouflaged Object Detection.}} Table \uppercase\expandafter{\romannumeral1} presents the results of TS-SAM compared with SAM, SAM-Adapter and the SOTA domain-specific models on four commonly used COD datasets.
With only fine-tuning 4.4\% of the parameters and without designing specific modules for this task, our model achieves competitive performance on all four datasets.
It is particularly notable on the two largest datasets, COD10K and NC4K, where TS-SAM achieves the best results on most metrics.
For the COD10K dataset, our model outperforms the second-best method, SARNet, by 3.3\% and 4.4\% in $S_{\alpha}$ and $F_{\beta}^{w}$, respectively, and reduced the $MAE$ by 19.0\%.
On the NC4K dataset, our model achieves SOTA performance in $S_{\alpha}$, $F_{\beta}^{w}$ and $MAE$, and surpassed SARNet by 1.8\% in $S_{\alpha}$.
These results demonstrate the robust generalization capability of our model.\par

Besides, for SAM-Adapter\cite{c4}, by introducing the adapter, it achieves the performance improvements for the COD task over the original SAM.
While for TS-SAM, by integrating the Convolutional Side Adapter in series, incorporating SAM feature maps into the training of the Convolutional Side Adapter, and employing the Multi-Scale Refinement Module, it achieves significant improvements across all metrics on the three datasets reported by SAM-Adapter\cite{c4}.
%These results indicate that TS-SAM, extracting features from the SAM image encoder and adapting them to downstream tasks has a great advantage over SAM-Adapter.\par

Fig. \ref{fig:compare_on_cod} presents some qualitative results on complex samples, showcasing various aspects of complexity.
For instance, there are interferences caused by extremely similar environments and occlusions (columns 1, 3, and 5), tiny targets (column 4) and complex texture features (column 2).
These results demonstrate the superiority of our proposed TS-SAM, including a good grasp of overall features, in-depth exploration of detailed features, and effective discrimination of interfering information.\par

\noindent{\textbf{Shadow detection.} The results on the ISTD dataset are shown in Table \uppercase\expandafter{\romannumeral2}.
It is observed that SAM-Adapter, by introducing lightweight adapters, significantly improves SAM's performance in the shadow detection task.
Compared to SAM-Adapter, TS-SAM has a significant performance improvement.
And without special design for the shadow detection task, TS-SAM achieves performance close to that of the best domain-specific model.}\par
% \vspace{-2em}
\noindent{\textbf{Salient Object Detection.} Table \uppercase\expandafter{\romannumeral3}  presents the results of TS-SAM compared with SAM, SSOM\cite{c5} and the SOTA domain-specific models on five commonly used SOD datasets. Compared to the similar SAM-based method SSOM, TS-SAM showed a significant advantage, greatly surpassing SSOM in the $MAE$ metric across all five datasets. This indicates that our proposed fine-tuning approach is superior to the AdaLora-based\cite{c15} fine-tuning scheme proposed by SSOM. Compared to advanced domain-specific methods, TS-SAM still achieves competitive performance with only fine-tuning a small number of parameters. For the $S_\alpha$, $E_\phi$, and $MAE$ metrics, TS-SAM reaching SOTA performance on the ECSSD, OMRON, and PASCAL-S datasets, and also performance well on the DUTS and HKU-IS datasets. For the $F_\beta^\omega$ metric, TS-SAM's performance is relatively weaker, which will be a direction for our future improvements.\par

% Overall, TS-SAM significantly enhances the performance of SAM in three tasks: COD, SOD and shadow detection, achieving a comparable level to the current SOTA models. This demonstrates that our fine-tuning approach can significantly improve the generalization ability of SAM in various downstream tasks, providing insights for the wider application of SAM in a broader range of downstream tasks.
\begin{table*}[htbp]
\centering
\caption{Quantitative comparison of TS-SAM with SAM, SSOM and the SOTA domain-specific models on four datasets of the SOD task: DUTS, ECSSD, OMRON, HKU-IS and PASCAL-S. The best results are marked in \textbf{bold}.}\label{tab3}
\scalebox{0.69}{
\begin{tabular}{ccccccccccccccccccccc}
\hline
\multicolumn{1}{c||}{\multirow{2}{*}{Methods}} 
& \multicolumn{4}{c|}{\textbf{DUTS}}
& \multicolumn{4}{c|}{\textbf{ECSSD}}
& \multicolumn{4}{c|}{\textbf{OMRON}}
& \multicolumn{4}{c|}{\textbf{HKU-IS}}
& \multicolumn{4}{c}{\textbf{PASCAL-S}} \\ \cline{2-21}

\multicolumn{1}{c||}{}                
& $S_\alpha \uparrow$ & $E_\phi\uparrow$ 
    & $F_\beta^\omega\uparrow$ & \multicolumn{1}{c|}{$MAE\downarrow$}
& $S_\alpha \uparrow$ & $E_\phi\uparrow$ 
    & $F_\beta^\omega\uparrow$ & \multicolumn{1}{c|}{$MAE\downarrow$}
& $S_\alpha \uparrow$ & $E_\phi\uparrow$ 
    & $F_\beta^\omega\uparrow$ &\multicolumn{1}{c|}{$MAE\downarrow$}
& $S_\alpha \uparrow$ & $E_\phi\uparrow$ 
    & $F_\beta^\omega\uparrow$ &\multicolumn{1}{c|}{$MAE\downarrow$}
& $S_\alpha \uparrow$ & $E_\phi\uparrow$ 
    & $F_\beta^\omega\uparrow$ &\multicolumn{1}{c}{$MAE\downarrow$}
\\ \hline

\multicolumn{1}{c||}{$\rm ICON_{22}$\cite{icon}}
    & \multicolumn{1}{c}{\textbf{0.917}} & \multicolumn{1}{c}{0.930}
    & \multicolumn{1}{c}{\textbf{0.886}} & \multicolumn{1}{c|}{\textbf{0.025}}  
    & \multicolumn{1}{c}{0.941} & \multicolumn{1}{c}{0.932} 
    & \multicolumn{1}{c}{0.936} & \multicolumn{1}{c|}{0.023} 
    & \multicolumn{1}{c}{0.869} & \multicolumn{1}{c}{0.898} 
    & \multicolumn{1}{c}{\textbf{0.804}} & \multicolumn{1}{c|}{0.043} 
    & \multicolumn{1}{c}{0.935} & \multicolumn{1}{c}{\textbf{0.965}} 
    & \multicolumn{1}{c}{0.925} & \multicolumn{1}{c|}{0.022} 
    & \multicolumn{1}{c}{0.885} & \multicolumn{1}{c}{0.875} 
    & \multicolumn{1}{c}{0.860} & \multicolumn{1}{c}{0.048} 
    \\

\multicolumn{1}{c||}{$\rm MENet_{23}$\cite{menet}}
    & \multicolumn{1}{c}{0.905} & \multicolumn{1}{c}{0.921}
    & \multicolumn{1}{c}{0.870} & \multicolumn{1}{c|}{0.028}  
    & \multicolumn{1}{c}{0.928} & \multicolumn{1}{c}{0.925} 
    & \multicolumn{1}{c}{0.920} & \multicolumn{1}{c|}{0.031} 
    & \multicolumn{1}{c}{0.850} & \multicolumn{1}{c}{0.882} 
    & \multicolumn{1}{c}{0.771} & \multicolumn{1}{c|}{0.045} 
    & \multicolumn{1}{c}{0.927} & \multicolumn{1}{c}{0.960} 
    & \multicolumn{1}{c}{0.917} & \multicolumn{1}{c|}{0.023} 
    & \multicolumn{1}{c}{0.871} & \multicolumn{1}{c}{0.870} 
    & \multicolumn{1}{c}{0.844} & \multicolumn{1}{c}{0.054} 
    \\

\multicolumn{1}{c||}{$\rm BBRF_{23}$\cite{bbrf}}
    & \multicolumn{1}{c}{0.909} & \multicolumn{1}{c}{0.927}
    & \multicolumn{1}{c}{\textbf{0.886}} & \multicolumn{1}{c|}{\textbf{0.025}}  
    & \multicolumn{1}{c}{0.939} & \multicolumn{1}{c}{0.934} 
    & \multicolumn{1}{c}{\textbf{0.944}} & \multicolumn{1}{c|}{\textbf{0.022}} 
    & \multicolumn{1}{c}{0.861} & \multicolumn{1}{c}{0.891} 
    & \multicolumn{1}{c}{0.803} & \multicolumn{1}{c|}{0.044} 
    & \multicolumn{1}{c}{0.932} & \multicolumn{1}{c}{\textbf{0.965}} 
    & \multicolumn{1}{c}{\textbf{0.932}} & \multicolumn{1}{c|}{\textbf{0.020}} 
    & \multicolumn{1}{c}{0.878} & \multicolumn{1}{c}{0.873} 
    & \multicolumn{1}{c}{\textbf{0.862}} & \multicolumn{1}{c}{0.049} 
    \\

\multicolumn{1}{c||}{$\rm SelfReformer_{23}$\cite{selfreformer}}
    & \multicolumn{1}{c}{0.911} & \multicolumn{1}{c}{0.921}
    & \multicolumn{1}{c}{0.872} & \multicolumn{1}{c|}{0.027}  
    & \multicolumn{1}{c}{0.936} & \multicolumn{1}{c}{0.929} 
    & \multicolumn{1}{c}{0.926} & \multicolumn{1}{c|}{0.027} 
    & \multicolumn{1}{c}{0.861} & \multicolumn{1}{c}{0.889} 
    & \multicolumn{1}{c}{0.784} & \multicolumn{1}{c|}{0.043} 
    & \multicolumn{1}{c}{0.931} & \multicolumn{1}{c}{0.959} 
    & \multicolumn{1}{c}{0.915} & \multicolumn{1}{c|}{0.024} 
    & \multicolumn{1}{c}{0.881} & \multicolumn{1}{c}{0.879} 
    & \multicolumn{1}{c}{0.854} & \multicolumn{1}{c}{0.051} 
    \\ \hline
    
\multicolumn{1}{c||}{$\rm SAM_{23}$\cite{c3}} 
    & \multicolumn{1}{c}{0.878} & \multicolumn{1}{c}{0.907} 
    & \multicolumn{1}{c}{0.794} & \multicolumn{1}{c|}{0.043}  
    & \multicolumn{1}{c}{0.931} & \multicolumn{1}{c}{0.949} 
    & \multicolumn{1}{c}{0.890} & \multicolumn{1}{c|}{0.033} 
    & \multicolumn{1}{c}{0.848} & \multicolumn{1}{c}{0.872} 
    & \multicolumn{1}{c}{0.716} & \multicolumn{1}{c|}{0.051} 
    & \multicolumn{1}{c}{0.920} & \multicolumn{1}{c}{0.949} 
    & \multicolumn{1}{c}{0.877} & \multicolumn{1}{c|}{0.030}
    & \multicolumn{1}{c}{0.867} & \multicolumn{1}{c}{0.897} 
    & \multicolumn{1}{c}{0.793} & \multicolumn{1}{c}{0.064} 
    \\
    
\multicolumn{1}{c||}{$\rm SSOM_{23}$\cite{c5}}
    & \multicolumn{1}{c}{-} & \multicolumn{1}{c}{-} 
    & \multicolumn{1}{c}{-} & \multicolumn{1}{c|}{0.034}  
    & \multicolumn{1}{c}{-} & \multicolumn{1}{c}{-} 
    & \multicolumn{1}{c}{-} & \multicolumn{1}{c|}{0.029} 
    & \multicolumn{1}{c}{-} & \multicolumn{1}{c}{-} 
    & \multicolumn{1}{c}{-} & \multicolumn{1}{c|}{0.048} 
    & \multicolumn{1}{c}{-} & \multicolumn{1}{c}{-} 
    & \multicolumn{1}{c}{-} & \multicolumn{1}{c|}{0.027}
    & \multicolumn{1}{c}{-} & \multicolumn{1}{c}{-} 
    & \multicolumn{1}{c}{-} & \multicolumn{1}{c}{0.062} 
    \\

\multicolumn{1}{c||}{$\rm TS-SAM\_B(Ours)$} 
    & \multicolumn{1}{c}{0.895} & \multicolumn{1}{c}{0.925} 
    & \multicolumn{1}{c}{0.827} & \multicolumn{1}{c|}{0.036}  
    & \multicolumn{1}{c}{0.937} & \multicolumn{1}{c}{0.955} 
    & \multicolumn{1}{c}{0.905} & \multicolumn{1}{c|}{0.029} 
    & \multicolumn{1}{c}{0.864} & \multicolumn{1}{c}{0.864} 
    & \multicolumn{1}{c}{0.744} & \multicolumn{1}{c|}{0.048} 
    & \multicolumn{1}{c}{0.929} & \multicolumn{1}{c}{0.958} 
    & \multicolumn{1}{c}{0.886} & \multicolumn{1}{c|}{0.026}
    & \multicolumn{1}{c}{0.868} & \multicolumn{1}{c}{0.901} 
    & \multicolumn{1}{c}{0.794} & \multicolumn{1}{c}{0.062} 
    \\

\multicolumn{1}{c||}{$\rm TS-SAM\_H(Ours)$} 
    & \multicolumn{1}{c}{0.915} & \multicolumn{1}{c}{\textbf{0.942}} 
    & \multicolumn{1}{c}{0.876} & \multicolumn{1}{c|}{0.028}  
    & \multicolumn{1}{c}{\textbf{0.948}} & \multicolumn{1}{c}{\textbf{0.965}} 
    & \multicolumn{1}{c}{0.926} & \multicolumn{1}{c|}{\textbf{0.022}} 
    & \multicolumn{1}{c}{\textbf{0.879}} & \multicolumn{1}{c}{\textbf{0.899}} 
    & \multicolumn{1}{c}{0.788} & \multicolumn{1}{c|}{\textbf{0.040}} 
    & \multicolumn{1}{c}{\textbf{0.937}} & \multicolumn{1}{c}{0.964} 
    & \multicolumn{1}{c}{0.903} & \multicolumn{1}{c|}{\textbf{0.020}}
    & \multicolumn{1}{c}{\textbf{0.892}} & \multicolumn{1}{c}{\textbf{0.924}} 
    & \multicolumn{1}{c}{0.837} & \multicolumn{1}{c}{\textbf{0.047}} 
    
\\ \hline
\end{tabular}
}
\end{table*}
\begin{table*}[t]
\centering
\caption{Effectiveness of the components on three datasets. The best results are marked in \textbf{bold}.}\label{tab4}
\scalebox{0.8}{
    \begin{tabular}{ccccccccccccccccccc}
    \hline
    
    \multicolumn{2}{c|}{\textbf{Components}}
    & \multicolumn{1}{c|}{\textbf{Tunable}}
    & \multicolumn{4}{c|}{\textbf{CHAMELEON}}
    & \multicolumn{4}{c|}{\textbf{CAMO}}
    & \multicolumn{4}{c|}{\textbf{COD10K}}
    & \multicolumn{4}{c}{\textbf{NC4K}}\\ \cline{1-2} \cline{4-19}
                  
    \multicolumn{1}{c}{CSA} & \multicolumn{1}{c|}{MRM+FFD}
    & \multicolumn{1}{c|}{\textbf{Param}}  
    & \multicolumn{1}{c}{$S_\alpha \uparrow$} 
        & \multicolumn{1}{c}{$E_\phi\uparrow$} 
        & \multicolumn{1}{c}{$F_\beta^\omega\uparrow$} 
        & \multicolumn{1}{c|}{$MAE\downarrow$}
    & \multicolumn{1}{c}{$S_\alpha \uparrow$} 
        & \multicolumn{1}{c}{$E_\phi\uparrow$} 
        & \multicolumn{1}{c}{$F_\beta^\omega\uparrow$} 
        & \multicolumn{1}{c|}{$MAE\downarrow$}
    & \multicolumn{1}{c}{$S_\alpha \uparrow$} 
        & \multicolumn{1}{c}{$E_\phi\uparrow$} 
        & \multicolumn{1}{c}{$F_\beta^\omega\uparrow$} 
        & \multicolumn{1}{c|}{$MAE\downarrow$}
    & \multicolumn{1}{c}{$S_\alpha \uparrow$} 
        & \multicolumn{1}{c}{$E_\phi\uparrow$} 
        & \multicolumn{1}{c}{$F_\beta^\omega\uparrow$} 
        & \multicolumn{1}{c}{$MAE\downarrow$}
    
    \\ \hline
    \multicolumn{2}{c|}{} & \multicolumn{1}{c|}{4.06M}
        & \multicolumn{1}{c}{0.786} & \multicolumn{1}{c}{0.802}
        & \multicolumn{1}{c}{0.610} & \multicolumn{1}{c|}{0.078}
        & \multicolumn{1}{c}{0.743} & \multicolumn{1}{c}{0.763}
        & \multicolumn{1}{c}{0.577} & \multicolumn{1}{c|}{0.121}
        & \multicolumn{1}{c}{0.767} & \multicolumn{1}{c}{0.806}
        & \multicolumn{1}{c}{0.555} & \multicolumn{1}{c|}{0.060} 
        & \multicolumn{1}{c}{0.786} & \multicolumn{1}{c}{0.807}
        & \multicolumn{1}{c}{0.630} & \multicolumn{1}{c}{0.084}\\
    
    \multicolumn{1}{c}{\Checkmark} & \multicolumn{1}{c|}{}
        & \multicolumn{1}{c|}{9.55M}
        & \multicolumn{1}{c}{0.903} & \multicolumn{1}{c}{0.936}
        & \multicolumn{1}{c}{0.834} & \multicolumn{1}{c|}{0.026}
        & \multicolumn{1}{c}{0.817} & \multicolumn{1}{c}{0.857}
        & \multicolumn{1}{c}{0.741} & \multicolumn{1}{c|}{0.077}
        & \multicolumn{1}{c}{0.857} & \multicolumn{1}{c}{0.906}
        & \multicolumn{1}{c}{0.758} & \multicolumn{1}{c|}{0.031}
        & \multicolumn{1}{c}{0.863} & \multicolumn{1}{c}{0.899}
        & \multicolumn{1}{c}{0.795} & \multicolumn{1}{c}{0.046}\\
    
    \multicolumn{1}{c}{} & \multicolumn{1}{c|}{\Checkmark}
        & \multicolumn{1}{c|}{6.41M}
        & \multicolumn{1}{c}{0.844} & \multicolumn{1}{c}{0.875}
        & \multicolumn{1}{c}{0.702} & \multicolumn{1}{c|}{0.047}
        & \multicolumn{1}{c}{0.772} & \multicolumn{1}{c}{0.811}
        & \multicolumn{1}{c}{0.655} & \multicolumn{1}{c|}{0.097}
        & \multicolumn{1}{c}{0.811} & \multicolumn{1}{c}{0.858}
        & \multicolumn{1}{c}{0.653} & \multicolumn{1}{c|}{0.040}
        & \multicolumn{1}{c}{0.816} & \multicolumn{1}{c}{0.854}
        & \multicolumn{1}{c}{0.688} & \multicolumn{1}{c}{0.064}\\
    
    \multicolumn{1}{c}{\Checkmark} & \multicolumn{1}{c|}{\Checkmark}
        & \multicolumn{1}{c|}{11.89M}
        & \multicolumn{1}{c}{\textbf{0.912}} 
        & \multicolumn{1}{c}{\textbf{0.947}}
        & \multicolumn{1}{c}{\textbf{0.849}} 
        & \multicolumn{1}{c|}{\textbf{0.023}}
        & \multicolumn{1}{c}{\textbf{0.826}} 
        & \multicolumn{1}{c}{\textbf{0.862}}
        & \multicolumn{1}{c}{\textbf{0.753}} 
        & \multicolumn{1}{c|}{\textbf{0.073}}
        & \multicolumn{1}{c}{\textbf{0.863}} 
        & \multicolumn{1}{c}{\textbf{0.909}}
        & \multicolumn{1}{c}{\textbf{0.771}} 
        & \multicolumn{1}{c|}{\textbf{0.029}}
        & \multicolumn{1}{c}{\textbf{0.866}} 
        & \multicolumn{1}{c}{\textbf{0.900}}
        & \multicolumn{1}{c}{\textbf{0.799}} 
        & \multicolumn{1}{c}{\textbf{0.045}}
    
    \\ \hline
    \end{tabular}
}
\end{table*}

\subsection{Ablation Study}
The effectiveness of our proposed modules was studied on four camouflaged object detection datasets: CHAMELEON, CAMO, COD10K and NC4K.
The modules used in TS-SAM were individually implemented on the baseline model of SAM, with results shown in Table \uppercase\expandafter{\romannumeral4}.
As seen from the results, the baseline model of SAM performs poor when only the decoder is fine-tuned.
First, we verify the effectiveness of the modules working independently, including settings where only the CSA module is introduced and where both the MRM and FFD modules were introduced.
It is found that compared to the baseline model, both settings bring significant performance improvements while adding only a small number of trainable parameters (5.49M and 2.35M, respectively). The results indicate that our proposed modules can effectively extract visual information from the SAM image encoder and adapt it to downstream tasks.
Second, the model that combines all three proposed modules achieves the best performance across all metrics on the four datasets.
The results indicate that our proposed modules can work in a complementary manner.
The MRM and FFD, designed specifically for segmentation tasks, can assist CSA in achieving better performance.\par

\section{Conclusion}
\label{sec:conclusion}
In this paper, we focus on the efficient fine-tuning of the large vision model SAM for downstream tasks.
To fully leverage the advantages of SAM pretrained on large-scale datasets, we introduce TS-SAM.
Using a lightweight Convolutional Side Adapter (CSA), we apply the concept of a side network to the fine-tuning of SAM for the first time.
Furthermore, in line with the characteristics of segmentation tasks, we designed the Multi-Scale Refinement Module (MRM) and the Feature Fusion Decoder (FFD) to extract the detailed features from the high-resolution image.
Experiments on three downstream tasks demonstrate that our model surpasses existing efficient fine-tuning methods for SAM, and it can achieve competitive performance compared to the SOTA domain-specific models specially designed for each task.
\section{Acknowledgement}
This work is partially supported by the National Natural Science Foundation (62272248), the Natural Science Foundation of Tianjin (21JCZDJC00740, 21JCYBJC00760, 23JCQNJC00010)

\bibliographystyle{IEEEbib}
\bibliography{reference}

\begin{thebibliography}{10}
\providecommand{\url}[1]{#1}
\csname url@samestyle\endcsname
\providecommand{\newblock}{\relax}
\providecommand{\bibinfo}[2]{#2}
\providecommand{\BIBentrySTDinterwordspacing}{\spaceskip=0pt\relax}
\providecommand{\BIBentryALTinterwordstretchfactor}{4}
\providecommand{\BIBentryALTinterwordspacing}{\spaceskip=\fontdimen2\font plus
\BIBentryALTinterwordstretchfactor\fontdimen3\font minus \fontdimen4\font\relax}
\providecommand{\BIBforeignlanguage}[2]{{%
\expandafter\ifx\csname l@#1\endcsname\relax
\typeout{** WARNING: IEEEtran.bst: No hyphenation pattern has been}%
\typeout{** loaded for the language `#1'. Using the pattern for}%
\typeout{** the default language instead.}%
\else
\language=\csname l@#1\endcsname
\fi
#2}}
\providecommand{\BIBdecl}{\relax}
\BIBdecl

\bibitem{c3}
A.~Kirillov, E.~Mintun, N.~Ravi, and et~al., ``Segment anything,'' \emph{arXiv preprint arXiv:2304.02643}, 2023.

\bibitem{c17}
L.~Tang, H.~Xiao, and B.~Li, ``Can sam segment anything? when sam meets camouflaged object detection,'' \emph{arXiv preprint arXiv:2304.04709}, 2023.

\bibitem{c4}
T.~Chen, L.~Zhu, C.~Deng, and et~al., ``Sam-adapter: Adapting segment anything in underperformed scenes,'' in \emph{ICCVW}, 2023, pp. 3367--3375.

\bibitem{c5}
\BIBentryALTinterwordspacing
Q.~Zhang, M.~Chen, A.~Bukharin, and et~al., ``Adaptive budget allocation for parameter-efficient fine-tuning,'' in \emph{ICLR}, 2023. [Online]. Available: \url{https://openreview.net/forum?id=lq62uWRJjiY}
\BIBentrySTDinterwordspacing

\bibitem{c8}
Y.-L. Sung, J.~Cho, and M.~Bansal, ``Lst: Ladder side-tuning for parameter and memory efficient transfer learning,'' \emph{NIPS}, vol.~35, pp. 12\,991--13\,005, 2022.

\bibitem{c9}
X.~L. Li and P.~Liang, ``Prefix-tuning: Optimizing continuous prompts for generation,'' \emph{arXiv preprint arXiv:2101.00190}, 2021.

\bibitem{c11}
M.~Jia, L.~Tang, B.-C. Chen, and et~al., ``Visual prompt tuning,'' in \emph{ECCV}.\hskip 1em plus 0.5em minus 0.4em\relax Springer, 2022, pp. 709--727.

\bibitem{c14}
M.~Xu, Z.~Zhang, F.~Wei, and et~al., ``Side adapter network for open-vocabulary semantic segmentation,'' in \emph{CVPR}, 2023, pp. 2945--2954.

\bibitem{c16}
W.~Lin, Z.~Wu, J.~Chen, and et~al., ``Hierarchical side-tuning for vision transformers,'' \emph{arXiv preprint arXiv:2310.05393}, 2023.

\bibitem{c20}
Z.~Yang, J.~Wang, Y.~Tang, and et~al., ``Lavt: Language-aware vision transformer for referring image segmentation,'' in \emph{CVPR}, 2022, pp. 18\,155--18\,165.

\bibitem{fdcod}
Y.~Zhong, B.~Li, L.~Tang, and et~al., ``Detecting camouflaged object in frequency domain,'' in \emph{CVPR}, 2022, pp. 4504--4513.

\bibitem{segmar}
Q.~Jia, S.~Yao, Y.~Liu, and et~al., ``Segment, magnify and reiterate: Detecting camouflaged objects the hard way,'' in \emph{CVPR}, 2022, pp. 4713--4722.

\bibitem{zoomnet}
Y.~Pang, X.~Zhao, T.-Z. Xiang, and et~al., ``Zoom in and out: A mixed-scale triplet network for camouflaged object detection,'' in \emph{CVPR}, 2022, pp. 2160--2170.

\bibitem{sinetv2}
D.-P. Fan, G.-P. Ji, M.-M. Cheng, and et~al., ``Concealed object detection,'' \emph{TPAMI}, vol.~44, no.~10, pp. 6024--6042, 2021.

\bibitem{fapnet}
T.~Zhou, Y.~Zhou, C.~Gong, and et~al., ``Feature aggregation and propagation network for camouflaged object detection,'' \emph{TIP}, vol.~31, pp. 7036--7047, 2022.

\bibitem{icon}
M.~Zhuge, D.-P. Fan, N.~Liu, D.~Zhang, D.~Xu, and L.~Shao, ``Salient object detection via integrity learning,'' \emph{TPAMI}, 2022.

\bibitem{feder}
C.~He, K.~Li, Y.~Zhang, and et~al., ``Camouflaged object detection with feature decomposition and edge reconstruction,'' in \emph{CVPR}, 2023, pp. 22\,046--22\,055.

\bibitem{dgnet}
G.-P. Ji, D.-P. Fan, Y.-C. Chou, and et~al., ``Deep gradient learning for efficient camouflaged object detection,'' \emph{Machine Intelligence Research}, vol.~20, no.~1, pp. 92--108, 2023.

\bibitem{sarnet}
H.~Xing, Y.~Wang, X.~Wei, and et~al., ``Go closer to see better: Camouflaged object detection via object area amplification and figure-ground conversion,'' \emph{TCSVT}, 2023.

\bibitem{fspnet}
Z.~Huang, H.~Dai, T.-Z. Xiang, and et~al., ``Feature shrinkage pyramid for camouflaged object detection with transformers,'' in \emph{CVPR}, 2023, pp. 5557--5566.

\bibitem{mscafnet}
Y.~Liu, H.~Li, J.~Cheng, and et~al., ``Mscaf-net: a general framework for camouflaged object detection via learning multi-scale context-aware features,'' \emph{TCSVT}, 2023.

\bibitem{hitnet}
X.~Hu, S.~Wang, X.~Qin, and et~al., ``High-resolution iterative feedback network for camouflaged object detection,'' in \emph{AAAI}, vol.~37, no.~1, 2023, pp. 881--889.

\bibitem{camo}
T.-N. Le, T.~V. Nguyen, Z.~Nie, and et~al., ``Anabranch network for camouflaged object segmentation,'' \emph{Journal of Computer Vision and Image Understanding}, vol. 184, pp. 45--56, 2019.

\bibitem{cod10k1}
D.-P. Fan, G.-P. Ji, M.-M. Cheng, and et~al., ``Concealed object detection,'' \emph{TPAMI}, 2022.

\bibitem{chameleon}
P.~Skurowski, H.~Abdulameer, J.~B{\l}aszczyk, and et~al., ``Animal camouflage analysis: Chameleon database,'' \emph{Unpublished manuscript}, vol.~2, no.~6, p.~7, 2018.

\bibitem{nc4k}
Y.~Lyu, J.~Zhang, Y.~Dai, and et~al., ``Simultaneously localize, segment and rank the camouflaged objects,'' in \emph{CVPR}, 2021.

\bibitem{istd}
J.~Wang, X.~Li, and J.~Yang, ``Stacked conditional generative adversarial networks for jointly learning shadow detection and shadow removal,'' in \emph{CVPR}, 2018.

\bibitem{duts}
L.~Wang, H.~Lu, Y.~Wang, and et~al., ``Learning to detect salient objects with image-level supervision,'' in \emph{CVPR}, 2017.

\bibitem{ecssd}
R.~Tran, D.~Patrick, M.~Geyer, and et~al., ``Sad: Saliency-based defenses against adversarial examples,'' \emph{arXiv preprint arXiv:2003.04820}, 2020.

\bibitem{omron}
Z.~Chen, Q.~Xu, R.~Cong, and et~al., ``Global context-aware progressive aggregation network for salient object detection,'' in \emph{AAAI}, vol.~34, no.~07, 2020, pp. 10\,599--10\,606.

\bibitem{hkuis}
G.~Li and Y.~Yu, ``Deep contrast learning for salient object detection,'' in \emph{CVPR}, 2016, pp. 478--487.

\bibitem{pascals}
Y.~Xu, D.~Xu, X.~Hong, and et~al., ``Structured modeling of joint deep feature and prediction refinement for salient object detection,'' in \emph{ICCV}, 2019, pp. 3789--3798.

\bibitem{c21}
P.~Sun, W.~Zhang, H.~Wang, and et~al., ``Deep rgb-d saliency detection with depth-sensitive attention and automatic multi-modal fusion,'' in \emph{CVPR}, 2021, pp. 1407--1417.

\bibitem{sdcm}
Y.~Zhu, X.~Fu, C.~Cao, and et~al., ``Single image shadow detection via complementary mechanism,'' in \emph{ACM MM}, 2022, pp. 6717--6726.

\bibitem{transshadow}
L.~Jie and H.~Zhang, ``A fast and efficient network for single image shadow detection,'' in \emph{ICASSP}.\hskip 1em plus 0.5em minus 0.4em\relax IEEE, 2022, pp. 2634--2638.

\bibitem{fcsdnet}
J.~M.~J. Valanarasu and V.~M. Patel, ``Fine-context shadow detection using shadow removal,'' in \emph{WACV}, 2023, pp. 1705--1714.

\bibitem{rmlanet}
L.~Jie and H.~Zhang, ``Rmlanet: Random multi-level attention network for shadow detection and removal,'' \emph{TCSVT}, 2023.

\bibitem{sddnet}
R.~Cong, Y.~Guan, J.~Chen, and et~al., ``Sddnet: Style-guided dual-layer disentanglement network for shadow detection,'' in \emph{ACM MM}, 2023, pp. 1202--1211.

\bibitem{sara}
J.~Sun, K.~Xu, Y.~Pang, and et~al., ``Adaptive illumination mapping for shadow detection in raw images,'' in \emph{ICCV}, 2023, pp. 12\,709--12\,718.

\bibitem{c15}
Q.~Zhang, M.~Chen, A.~Bukharin, and et~al., ``Adaptive budget allocation for parameter-efficient fine-tuning,'' \emph{arXiv preprint arXiv:2303.10512}, 2023.

\bibitem{menet}
Y.~Wang, R.~Wang, X.~Fan, and et~al., ``Pixels, regions, and objects: Multiple enhancement for salient object detection,'' in \emph{CVPR}, 2023, pp. 10\,031--10\,040.

\bibitem{bbrf}
M.~Ma, C.~Xia, C.~Xie, and et~al., ``Boosting broader receptive fields for salient object detection,'' \emph{TIP}, vol.~32, pp. 1026--1038, 2023.

\bibitem{selfreformer}
Y.~K. Yun and W.~Lin, ``Towards a complete and detail-preserved salient object detection,'' \emph{TMM}, 2023.

\end{thebibliography}

\end{document}